\pdfoutput=1
\documentclass{bmvc2k}

\title{Divide and Fuse: A Re-ranking Approach for Person Re-identification}

\addauthor{Rui Yu}{yurui.thu@gmail.com}{1}
\addauthor{Zhichao Zhou}{zzc@hust.edu.cn}{1}
\addauthor{Song Bai}{songbai@hust.edu.cn}{1}
\addauthor{Xiang Bai $^{\ast}$}{xbai@hust.edu.cn}{1}

\addinstitution{
 Huazhong University of Science and Technology\\
 Wuhan, China
}

\runninghead{Yu et al.}{Divide and Fuse: a Re-ranking Approach for Person Re-ID}

\def\eg{\emph{e.g}\bmvaOneDot}

\def\etal{\emph{et al}\bmvaOneDot}

\usepackage{multirow}
\usepackage[symbol]{footmisc}
\begin{document}

\maketitle

\begin{abstract}
As re-ranking is a necessary procedure to boost person re-identification (re-ID) performance on large-scale datasets, the diversity of feature becomes crucial to person re-ID for its importance both on designing pedestrian descriptions and re-ranking based on feature fusion. However, in many circumstances, only one type of pedestrian feature is available. In this paper, we propose a ``Divide and Fuse'' re-ranking framework for person re-ID. It exploits the diversity from different parts of a high-dimensional feature vector for fusion-based re-ranking, while no other features are accessible. Specifically, given an image, the extracted feature is divided into sub-features. Then the contextual information of each sub-feature is iteratively encoded into a new feature. Finally, the new features from the same image are fused into one vector for re-ranking. Experimental results on two person re-ID benchmarks demonstrate the effectiveness of the proposed framework. Especially, our method outperforms the state-of-the-art on the Market-1501 dataset.
\end{abstract}

\section{Introduction}
\label{sec:intro}
\begin{figure}[!tb]
\centering
\includegraphics[width=1\textwidth]{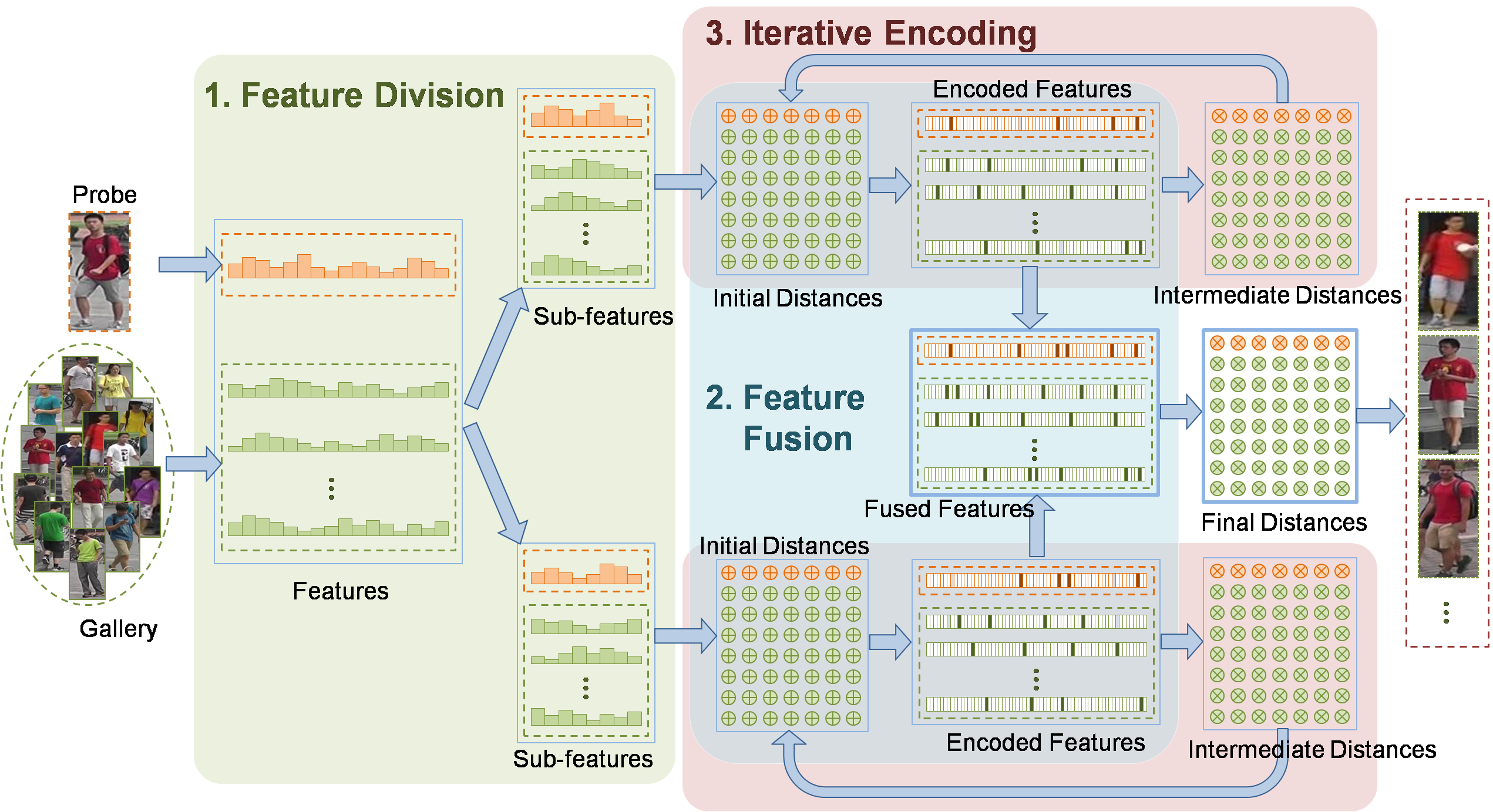}
\caption{The proposed ``Divide and Fuse'' re-ranking framework in the case of ``$L=2$''. $L$ represents the number of divided parts. We set $L$ around 10 in the experiments.}
\label{fig:pipeline}
\end{figure}

Person re-identification~\cite{zheng2016person,bedagkar2014survey}, which aims to retrieve the same identity from the gallery set for a given query image (the probe), has recently drawn increasing attention from both academia and industry due to its important applications in intelligent surveillance. It remains a rather challenging task owing to large variations in pose, viewpoint, illumination and occlusion.

Current research interests in person re-ID mainly focus on two aspects: 1) extracting visual features~\cite{cheng2011custom,ma2012bicov,yang2014salient,liao2015person,zheng2015query,matsukawa2016hierarchical} that preserve the appearance information of a person; 2) learning a discriminative metric~\cite{hirzer2012relaxed,zheng2013reidentification,li2013learning,xiong2014person,chen2015similarity,liao2015efficient} which minimizes the distance between features from the same identity. The ranking list is built based on the distances between the probe and gallery features. Note that person re-ID is similar to instance retrieval in the testing phase, thus, re-ranking (\eg ~\cite{bai2017scalable}) has attracted much attention gradually, which could boost the re-ID performance. In this paper, we mainly work on devising an effective and efficient re-ranking framework to improve the re-ID accuracy.

When designing robust hand-crafted pedestrian descriptions, different types of features are usually incorporated to expand the diversity. For example, Gray and Tao~\cite{gray2008viewpoint} utilize 8 color channels (RGB, HS, and YCbCr) and 21 texture filters on the luminance channel to build the feature vector. Instead of simply concatenating the diversified features, the fusion-based re-ranking can result in better performance. Multiple feature fusion has been proven effective in re-ranking algorithms~\cite{zhang2015query}. However, superior multiple features are not always easily accessible. The features extracted from the deep neutral network generally outperform hand-crafted alternatives on large-scale person re-ID benchmarks~\cite{zheng2015scalable,zheng2016mars}. Despite the fact that many deep-learning-based approaches~\cite{varior2016gated,li2014deepreid,Cheng_2016_CVPR,NIPS2016_6367} have been proposed for person re-ID, it would be burdensome and time-consuming to obtain multiple deep features by training different networks. Then we cannot easily benefit from the fusion of multiple complementary features to promote re-ranking.

Our concern is how to fully exploit the information involved in a single feature to improve the retrieval accuracy, while no other features are available. Considering that the dimension of the extracted feature is typically high (\eg, 2,048-dim for CNN feature), the characteristics among different local parts of the feature can be rather diverse. This inspires us to broach a simple idea to bring in diversity for fusion: splitting the feature into parts rather than treating it as a whole. Based on this consideration, we present a ``Divide and Fuse'' (DaF) re-ranking framework for person re-ID as illustrated in Figure \ref{fig:pipeline}.
\begin{enumerate}
\item \textbf{Feature Division.} The extracted feature for each image is split into a certain number of parts, namely sub-features.
\item \textbf{Feature Fusion.} The contextual information of each sub-feature is encoded into a new feature. Then the new features encoded from each image are fused into a single vector. The final ranks are decided by the generalized Jaccard distances between the fused vectors.
\item \textbf{Iterative Encoding.} An iterative strategy is introduced to the feature encoding procedure to further improve re-ID performance.
\end{enumerate}

The contribution of our work is three-fold. 1) We present a feature ``Divide and Fuse'' approach for re-ranking person re-ID. It enables us to optimize the ranking list by fully utilizing both the diversity within a single feature and the contextual relations among different features. 2) The iterative feature encoding and fusion scheme can be directly applied to multiple feature fusion for re-ranking in a totally unsupervised manner. 3) Our re-ranking approach remarkably improves re-ID performances on multiple datasets, and in particular, outperforms the state-of-the-art on the popular Market-1501 dataset.

\section{Related Work}
\label{sec:related}
The re-ranking technique~\cite{mei2014multimedia} is generally used as a post-processing step in various retrieval problems, where the initial ranking list for a query is sorted based on the pairwise similarities between the query and the instances in the database. The re-ranking procedure refines the initial ranking list by taking account of the neighborhood relations among all the instances. A variety of re-ranking algorithms have been developed for object retrieval. In particular, Sparse Contextual Activation (SCA)~\cite{bai2016sparse} encodes the neighborhood set into a sparse vector and measures the sample dissimilarity in generalized Jaccard distance. Bai \etal ~\cite{Bai_AAAI17} provide theoretical explanations for diffusion process, which is a popular branch of re-ranking.

Though current research interests in person re-ID mainly rest on extracting robust feature representations and learning discriminative distance metrics, re-ranking techniques~\cite{li2012common,leng2015person,garcia2015person,ye2016person,zhong2017re,bai2017scalable} are drawing more and more attention in this field recently. Leng~\etal~\cite{leng2015person} propose a bidirectional re-ranking technique by reversely querying every gallery image in a new gallery composed of the probe and other gallery images. Discriminant context information analysis (DCIA)~\cite{garcia2015person} is an unsupervised post-ranking framework which analyzes the context information of the first ranks and removes the visual ambiguities. Ye~\etal~\cite{ye2016person} propose a ranking aggregation algorithm by using both similarity and dissimilarity cues from different baseline methods. More recently, as the counterpart and the application of SCA~\cite{bai2016sparse} in re-ID, Zhong~\etal~\cite{zhong2017re} introduce $k$-reciprocal encoding to re-ranking the gallery images. Bai \etal \cite{bai2017scalable} propose Supervised Smoothed Manifold (SSM) to estimate the similarity between two instances in the context of other pairs of instances. As a manifold-based affinity learning algorithm, SSM can boost performances of most existing methods as a post-ranking tool. Besides the proposed iterative feature encoding scheme, our re-ranking framework can be obviously distinguished from other methods that we pack the feature ``division and fusion'' strategy into the re-ranking process so as to gain performance improvement from the variety rooted in a single feature.
\section{Proposed Method}
\label{sec:method}

\subsection{Feature Division}
\label{subsec:division}
\begin{figure}[!tb]
\centering
\includegraphics[width=0.9\textwidth]{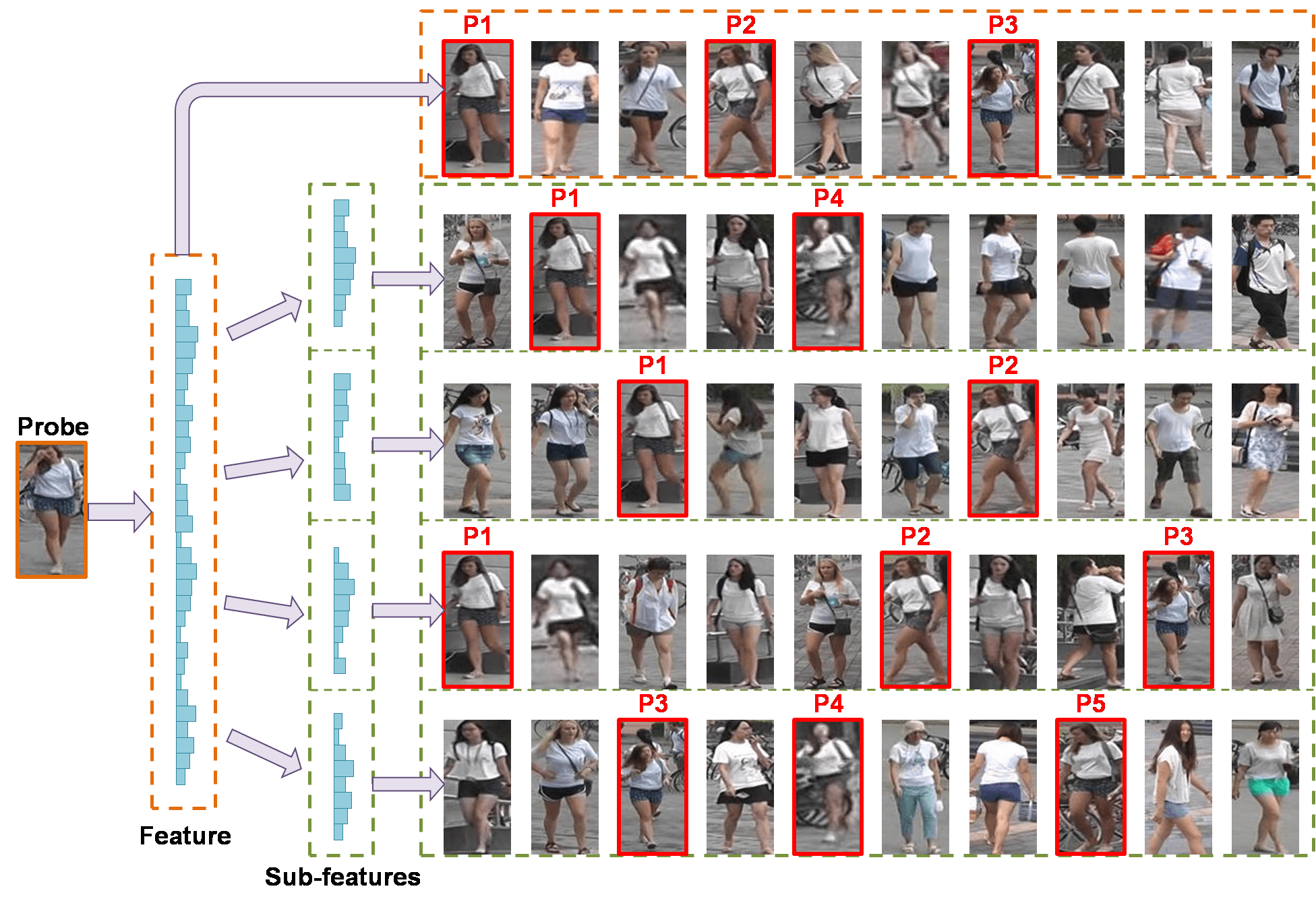}
\caption{Illustration of feature division for re-ranking. Thanks to the diversity among sub-features, more ground truths (P4, P5) are included in the top-ranks.}
\label{fig:divide}
\end{figure}

Suppose that the feature is approximately evenly divided into $L$ parts, namely sub-features. As long as $L$ is relatively small, each sub-feature would also be high-dimensional and therefore almost as informative as the global feature. While the information carried by the sub-features are diverse, aggregating the variety of sub-features may improve the overall discriminating power. For example, we divide the feature of a probe into four sub-features. Top-10 ranks of the global feature and each sub-feature are illustrated in Figure~\ref{fig:divide}. For the global feature, three ground truths (P1, P2, P3) are included in the top-10 ranks. While the ranking lists of four sub-features diverge from each other and bring two more ground truths (P4, P5) to the top ranks, the division provides possibilities of fusing the diverse information from different sub-features to improve the re-ID accuracy.

Note that the feature division scheme can be applied to both hand-crafted and deep-learned features. To split a feature, we can select elements manually, randomly, or even based on feature selection techniques. In this study, we directly adopt either a manual or a random manner for simplicity. For a CNN feature extracted from the fully-connected layer, every element in it is generated unbiasedly, so the way of selecting elements is theoretically of no importance. For a hand-crafted feature, the element selection manner depends on the structure of the feature. In general, we tend to construct the sub-features with more diversity among them.

Consider a probe image $p$ and the gallery set $G=\{g_i|i=1,2,\ldots,N_g\}$ with $N_g$ images. Let $x_p$ and $x_{g_i}$ denote the $M$-dimensional features extracted from the probe $p$ and a gallery $g_i$, respectively. As described above, we have each feature split into $L$ sub-features. The $l$-th sub-feature of $x_p$ and $x_{g_i}$ are denoted by $x^{(l)}_p$ and $x^{(l)}_{g_i}$, respectively. The initial distance $d(x^{(l)}_p,x^{(l)}_{g_i})$ between $x^{(l)}_p$ and $x^{(l)}_{g_i}$ can be computed under a certain metric, \eg, Euclidean distance. To simplify the notation, we use $d^{(l)}(p,g_i)$ to replace $d(x^{(l)}_p,x^{(l)}_{g_i})$ below.

For each probe $p$, an initial ranking list can be obtained based on $d^{(l)}(p,g_i)$. The rank of a gallery $g_j$ $(j=1,2,\ldots,N_g)$ in the list is denoted by $R^{(l)}_p(g_j)$. Similarly, for each $g_i$ $(i=1,2,\ldots,N_g)$, we can rank the gallery set based on $d^{(l)}(g_i,g_j)$ and let $R^{(l)}_{g_i}(g_j)$ denote the rank of $g_j$ in each list. Obviously $R^{(l)}_{g_i}(g_i)=1$ in general.

\subsection{Feature Fusion}
\label{subsec:fusion}
\noindent\textbf{Feature encoding.}~Before fusing the divided sub-features, we must figure out what kind of information from the sub-features is valuable. As illustrated in Figure~\ref{fig:divide}, the ranking lists of different sub-features introduce contrasting correct positives, which could be useful for feature fusion. So we are mainly concerned with utilizing the divergent neighborhood relations of sub-features. Based on this point, we first adopt an effective re-ranking component to encode the neighborhood relation of each sub-feature $x^{(l)}_p$ $(l=1,2,\ldots,L)$ into a new $N_g$-dimensional feature $V^{(l)}_p$ and then fuse the $L$ new features into one. The component is proposed by sparse contextual activation~\cite{bai2016sparse} for visual re-ranking and then utilized by $k$-reciprocal encoding~\cite{zhong2017re} for person re-ID. Specifically, the $j$-th element of the encoded feature is defined as
\begin{equation}
\label{eq:vecdef}
V^{(l)}_p[j]=
  \begin{cases}
  S^{(l)}(p,g_j) & \text{if $R^{(l)}_p(g_j)\leq k$}\\
  0 & \text{otherwise}
  \end{cases},
\end{equation}
where $S^{(l)}(\cdot,\cdot)$ is the similarity function, which is computed directly from the initial distance in~\cite{bai2016sparse} and ~\cite{zhong2017re}. Instead, in order to take full advantage of the contextual information, we compute the similarity from the neighborhood relations rather than directly from the initial distance. Inspired by~\cite{shen2012object}, we measure the similarity between probe and gallery by the gallery's ranks in the ranking lists of probe and probe's $k$-nearest neighbors. Specifically, the similarity is defined as
\begin{equation}
\label{eq:simil}
S^{(l)}(p,g_j)=\frac{1}{R^{(l)}_p(g_j)}+\sum_{m:R^{(l)}_p(g_m)\leq k}\frac{1}{R^{(l)}_{g_m}(g_j)\left(1+R^{(l)}_p(g_m)\right)},
\end{equation}
where $k$ is defined the same as in Eq.~\eqref{eq:vecdef} and denoted by $k_1$. Similarly, we can encode a new feature $V^{(l)}_{g_i}$ for every $x^{(l)}_{g_i}$ $(i=1,2,\ldots,N_g)$ to facilitate the neighbor enhancement
\begin{equation}
\label{eq:enhance}
V^{(l)}_p=\frac{1}{1+k}\left(V^{(l)}_p+\sum_{i:R^{(l)}_p(g_i)\leq k}V^{(l)}_{g_i} \right),
\end{equation}
where $k$ is denoted by $k_2$ to be distinguished from the $k$ in Eq.~\eqref{eq:vecdef} and Eq.~\eqref{eq:simil}. In this way, the encoded feature can benefit from the neighborhood information of the probe's neighbors.

\vspace{+1ex}\noindent\textbf{Fuzzy fusion.}~As $k_1$ is set far less than $N_g$, the encoded features are sparse vectors. Our goal is to fuse the $L$ sparse vectors $V^{(l)}_p$ into one $V_p$ while maintaining the useful information. In that way, the fused vector $V_p$ would be more discriminative than every single vector $V^{(l)}_p$. Since no prior information is given to measure the discriminating power of each encoded feature, we formulate the fusion process in an unsupervised manner. Here we adopt the simple yet effective fuzzy aggregation operator in fuzzy theory
\begin{equation}
\label{eq:fusion}
V_p[j]=\left(\frac{1}{L}\sum^L_{l=1}\left(V^{(l)}_p[j]\right)^{\alpha}\right)^{\frac{1}{\alpha}},
\end{equation}
where $V_p[j]$ denotes the $j$-th element of the fused vector $V_p$, and $\alpha$ is a fusion exponent.
The fuzzy aggregation operator is a generic averaging function, which can adapt to different scales of elements within the encoded features. When lacking prior information, it provides a flexible tool to effectively fuse sparse vectors. Particularly, with changes in $\alpha$, the operator can be converted to various common mean functions, \eg, arithmetic mean or geometric mean. Moreover, this element-wise fusion method never breaks the neighborhood relations encoded in the new features. For the same reason, the element-wise generalized Jaccard distance is adopted to finally measure the distances between the probe and galleries
\begin{equation}
\label{eq:Jaccard}
\hat{d}(p,g_i)=d_J(V_p,V_{g_i})=1-\frac{\sum_{j=1}^{N_g}\min(V_p[j],V_{g_i}[j])}{\sum_{j=1}^{N_g}\max(V_p[j],V_{g_i}[j])}.
\end{equation}
Since $V_p$ and $V_{g_i}$ are sparse vectors with non-negative elements, $\sum_{j=1}^{N_g}\min(V_p[j],V_{g_i}[j])=0$ (thus $\hat{d}(p,g_i)=1$) for most $g_i$ in the gallery. This accords with the fact that most gallery images don't belong to the same person of the probe.

\subsection{Iterative Encoding}
\label{subsec:iterative}
As for the feature encoding presented in Section~\ref{subsec:fusion}, there is no iterative scheme introduced in previous methods~\cite{bai2016sparse,zhong2017re}. In fact, the retrieval accuracy usually declines in practice when iteratively using the new distance to perform feature encoding. In order to exploit $V^{(l)}_p$ recurrently, we compute an intermediate distance for $V^{(l)}_p$ in the same way as Eq.~\eqref{eq:Jaccard}
\begin{equation}
\label{eq:sub_Jaccard}
\hat{d}^{(l)}(p,g_i)=d_J(V^{(l)}_p,V^{(l)}_{g_i})=1-\frac{\sum_{j=1}^{N_g}\min(V^{(l)}_p[j],V^{(l)}_{g_i}[j])}{\sum_{j=1}^{N_g}\max(V^{(l)}_p[j],V^{(l)}_{g_i}[j])}.
\end{equation}
Similar to the discussion in Section~\ref{subsec:fusion}, the value of $\hat{d}^{(l)}(p,g_i)$ equals to 1 for most $g_i$. That means the distance information out of the close neighbors is wiped out during the encoding procedure. When directly replacing the initial distance with the intermediate distance, the number of neighbors available for effective encoding is limited by the previous iteration, so probably impairing the re-ranking performance.

However, we can still improve the retrieval accuracy in an iterative manner. Since $V^{(l)}_p$ just encodes the neighborhood relations given by the initial distance $d^{(l)}(p,g_i)$, according to Eq.~\eqref{eq:sub_Jaccard}, the intermediate distance $\hat{d}^{(l)}(p,g_i)$ only includes the neighborhood information. So the initial distance and the intermediate distance are complementary. When combining the two types of distances, the intermediate distances can activate promising elements in the initial distance matrix from the viewpoint of neighborhood. As shown in Figure~\ref{fig:pipeline}, we iteratively renew the initial distance by aggregating the intermediate distance into it. The aggregation function can be simply defined as
\begin{equation}
\label{eq:dist_concate}
d^{(l)}(p,g_i):=(1-\lambda)d^{(l)}(p,g_i)+\lambda\hat{d}^{(l)}(p,g_i),
\end{equation}
where $\lambda \in (0,1)$ denotes the aggregating factor. The aggregated distance could be a better baseline for feature encoding by adding the contextual information to the initial distance. As a result, the new $V^{(l)}_p$ encoded from the updated distance will accordingly gain more discriminative power and further improve the re-ID accuracy.

\section{Experiments}
\label{sec:experiment}
The proposed DaF re-ranking method is evaluated on two large-scale person re-ID benchmark datasets, including Market-1501~\cite{zheng2015scalable} and CUHK03~\cite{li2014deepreid}.
Two evaluation metrics, the cumulative matching characteristics (CMC) and mean average precision (mAP), are used. We conducted two-iteration feature encoding with the aggregating factor $\lambda=0.2$ on both datasets. In the fusion process, we set $\alpha$ to 0.5 throughout the experiments.

\subsection{Market-1501 Dataset}

\begin{table}
\begin{center}
\begin{tabular}{|c|c|c|}
\hline
 Re-ranking Method & Rank-1 & \ mAP  \quad \\
\hline\hline
 Baseline & 78.92 & 55.03 \\
 CDM~\cite{jegou2007contextual} & 79.81 & 56.73 \\
 $k$-NN~\cite{shen2012object} & 79.54 & 60.26 \\
 SCA~\cite{bai2016sparse} & 79.75 & 68.97 \\
 $k$-RE~\cite{zhong2017re} & 80.55 & 69.57 \\
 \bf{DaF (Ours)} & \bf{82.30} & \bf{72.42} \\
\hline
\end{tabular}
\end{center}
\caption{Comparison with baseline on Market-1501 dataset (single query)}
\label{tab:market_baseline}
\end{table}

\begin{table}
\begin{center}
\begin{tabular}{|l c|c|c|}
\hline
Method & Ref & Rank-1 & mAP\\
\hline\hline
SSDAL~\cite{su2016deep} & ECCV 2016 & 39.40 & 19.60 \\
TMA~\cite{martinel2016temporal} & ECCV 2016 & 47.92 & 22.31 \\
SCSP~\cite{chen2016similarity} & CVPR 2016 & 51.90 & 26.35 \\
Null~\cite{zhang2016learning} & CVPR 2016 & 61.02 & 35.68 \\
S-CNN~\cite{varior2016gated} & ECCV 2016 & 65.88 & 39.55 \\
CRAFT-MFA~\cite{chen2017person} & TPAMI 2017 & 71.80 & 45.50 \\
$k$-RE~\cite{zhong2017re} & CVPR 2017 & 77.11 & 63.63 \\
SSM~\cite{bai2017scalable} & CVPR 2017 & 82.21 & 68.80 \\
\hline\hline
\bf{DaF (Ours)} &  & \bf{82.30} & \bf{72.42} \\
\hline
\end{tabular}
\end{center}
\caption{Comparison with state-of-the-art on Market-1501 dataset (single query)}
\label{tab:market_sota}
\end{table}

Market-1501~\cite{zheng2015scalable} is the largest image-based person re-ID benchmark to date, which consists of 32,668 labeled images of 1,501 identities captured by six cameras. Within the dataset, 12,936 images of 751 identities are used for training and the rest (19,732 images of 750 other identities) are used for testing. In the testing part, 3,368 images of 750 identities are taken as the probe set. Following the standard protocol, we report the single-query evaluation results on this dataset. As for the parameters in our algorithm, we set $k_1$ to 20, $k_2$ to 4, and $L$ to 11.

\vspace{1ex}\noindent\textbf{Comparison with baseline.} Owing to a large quantity of training images, the approaches based on deep neural networks are widely used by previous works~\cite{su2016deep,varior2016gated,varior2016siamese} on this database. Following this trend, we extract the 2,048-dim ID-discriminative Embedding (IDE)  feature by fine-tuning pre-trained ResNet~\cite{he2016deep} model. Under the Euclidean metric, we obtain a baseline performance of 78.92\% in rank-1 accuracy and 55.03\% in mAP. Our re-ranking method is evaluated on this baseline. When testing a given probe, we do not use the information from other probe instances. It would violate the standard protocol if utilizing the contextual information of other probes to promote re-ranking. As presented in Table~\ref{tab:market_baseline}, our re-ranking approach gains an impressive increase of 17.39\% in mAP and 3.38\% in rank-1 accuracy. Moreover, contrastive experiments using the same baseline show that our re-ranking method outperforms other competitive re-ranking algorithms for image retrieval and person re-ID, including contextual dissimilarity measure (CDM)~\cite{jegou2007contextual}, $k$-NN re-ranking ($k$-NN)~\cite{shen2012object}, sparse contextual activation (SCA)~\cite{bai2016sparse} and $k$-reciprocal encoding ($k$-RE)~\cite{zhong2017re}.

\vspace{+1ex}\noindent\textbf{Comparison with state-of-the-art.}~We compare the proposed method with other state-of-the-art methods, as listed in Table~\ref{tab:market_sota}. The performance of our method is reported based on the ResNet feature under Euclidean metric. The previous state-of-the-art performance is achieved by supervised smoothed manifold (SSM)~\cite{bai2017scalable}. Bai \etal ~\cite{bai2017scalable} propose a novel affinity learning algorithm which can boost performances of most existing methods. Based on a more delicately trained ResNet model, the baseline performance of SSM~\cite{bai2017scalable} (mAP 61.12\%) is higher than ours (mAP 55.03\%). Nevertheless, our method benefits from the ``Divide and Fuse'' re-ranking framework and outperforms SSM with improvement of 3.62\% in mAP.

\subsection{CUHK03 Dataset}

\begin{table}
\begin{center}
\begin{tabular}{|c|c|c|c|c|}
\hline
\multirow {2}{*}{Re-ranking Method} &
\multicolumn {2}{c|}{Labeled} & \multicolumn {2}{c|}{Detected} \\
\cline{2-5}
 & Rank-1 & mAP & Rank-1 & mAP \\
\hline\hline
 Baseline & 22.2 & 21.0 & 21.3 & 19.7 \\
 CDM~\cite{jegou2007contextual} & 24.8 & 22.7 & 22.9 & 20.6 \\
 $k$-NN~\cite{shen2012object} & 24.7 & 24.2 & 24.3 & 22.9 \\
 SCA~\cite{bai2016sparse} & 24.5 & 27.6 & 24.7 & 26.6 \\
 $k$-RE~\cite{zhong2017re} & 26.6 & 28.9 & 24.9 & 27.3 \\
 \bf{DaF (Ours)} & \bf{27.5} & \bf{31.5} & \bf{26.4} & \bf{30.0} \\
\hline
\end{tabular}
\end{center}
\caption{Comparison with baseline on CUHK03 dataset under new training/testing protocol}
\label{tab:cuhk03_baseline}
\end{table}

CUHK03~\cite{li2014deepreid} consists of 13,164 images of 1,360 identities captured by six surveillance cameras. Each identity shows in two camera views and has 4.8 images on average in one view. Besides the manually labeled bounding boxes, the dataset also provides automatically detected bounding boxes by the deformable-part-model (DPM) detector. We report experimental results on both "labeled" and "detected" data.

In the conventional single-shot experimental setting, the dataset is divided into a training set of 1,160 identities and a testing set of 100 identities. The experiments are conducted with 20 random splits to obtain an average result. To facilitate the comparison with other re-ranking methods, we tend to have multiple ground truths in the gallery for each probe. Thus, in this work, we report the experimental results following the new training/testing protocol introduced by~\cite{zhong2017re}. In the new protocol, the CUHK03 dataset is split into a training set of 767 identities and a testing set of 730 identities. In testing, one image of each identity is randomly selected from each camera, while the remaining images in the testing set are employed as the gallery set.

We set $k_1$ to 12, $k_2$ to 4, and $L$ to 7 in this experiment. The proposed re-ranking method is evaluated using the 2,048-dim ResNet features. As shown in Table~\ref{tab:cuhk03_baseline}, our method achieves remarkable performance improvement and outperforms other competitive re-ranking algorithms. For example, it gains an increase of 5.3\% in rank-1 accuracy and 10.5\% in mAP on the "labeled" data.

\subsection{Parameter Analysis}

\begin{figure}[!tb]
\begin{center}
\begin{tabular}{cccc}
\bmvaHangBox{\includegraphics[width=2.85cm]{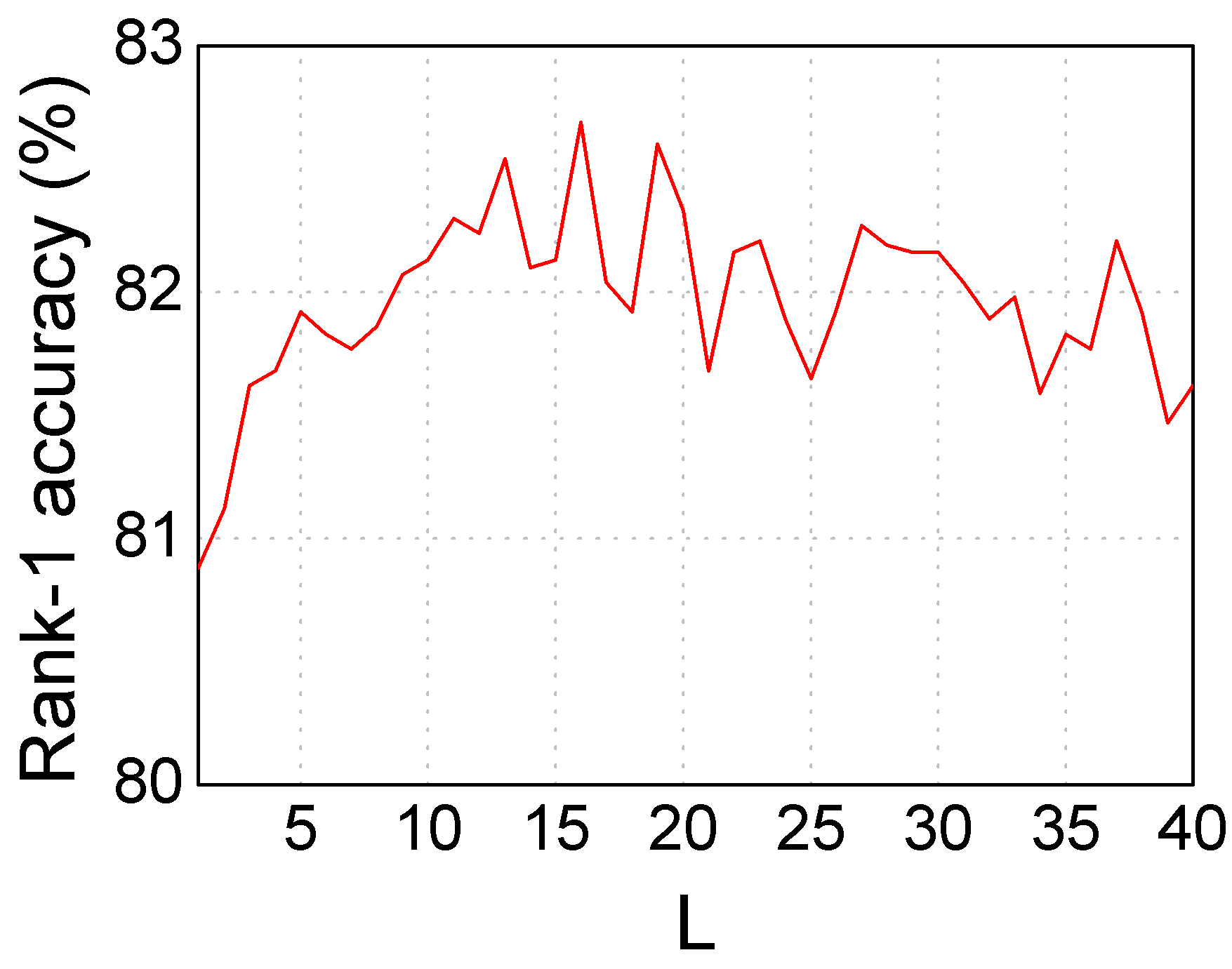}}&
\bmvaHangBox{\includegraphics[width=2.85cm]{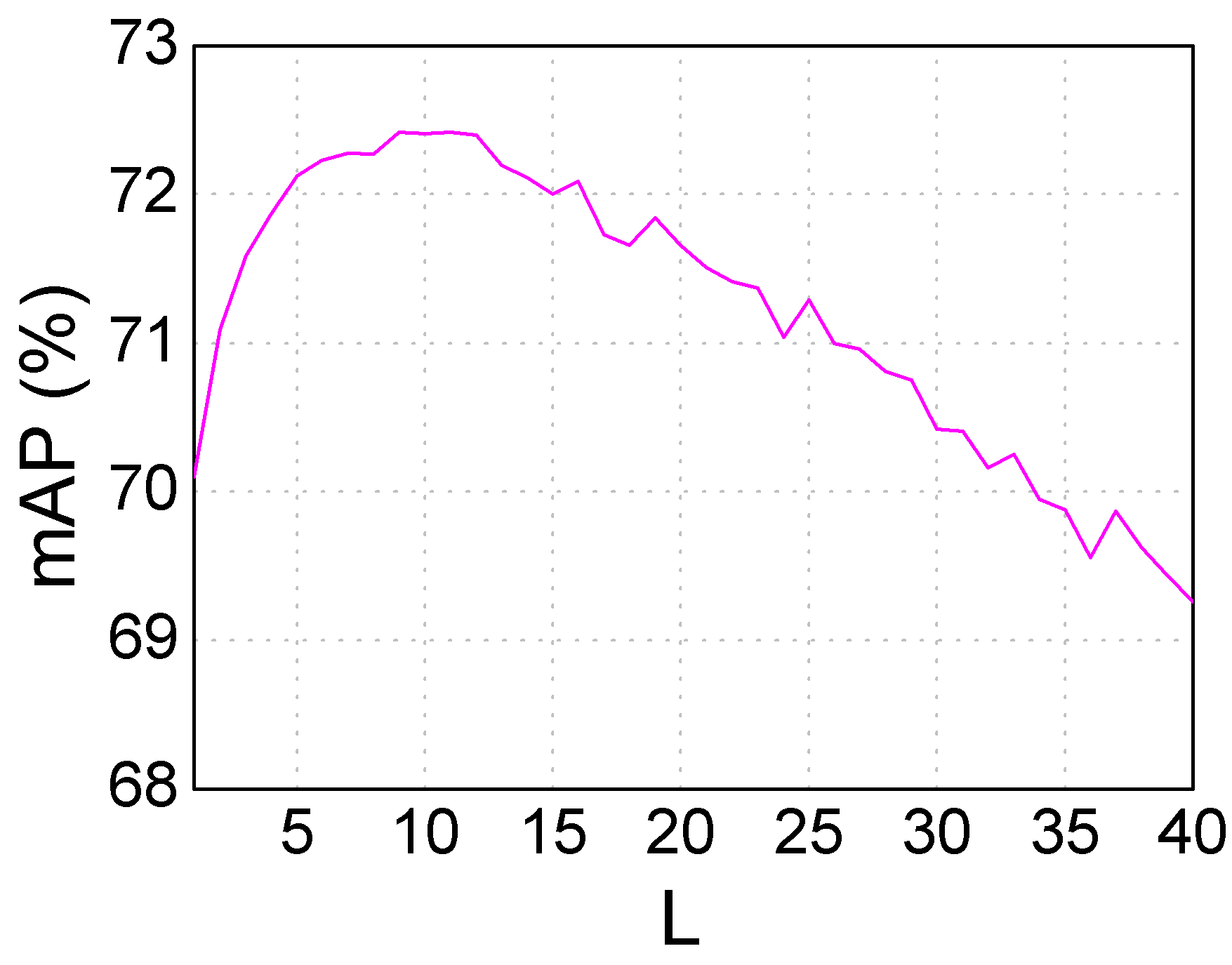}}&
\bmvaHangBox{\includegraphics[width=2.85cm]{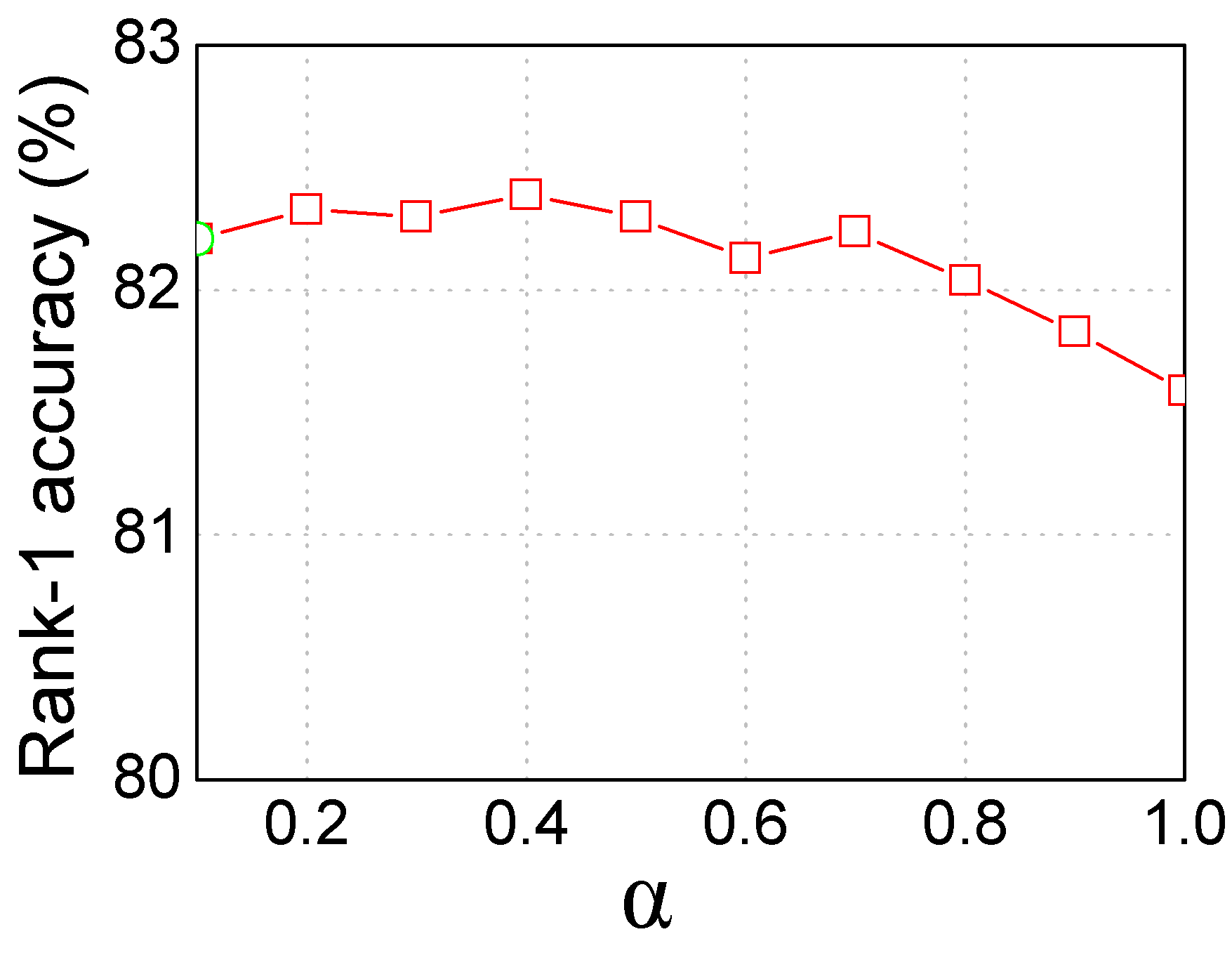}}&
\bmvaHangBox{\includegraphics[width=2.85cm]{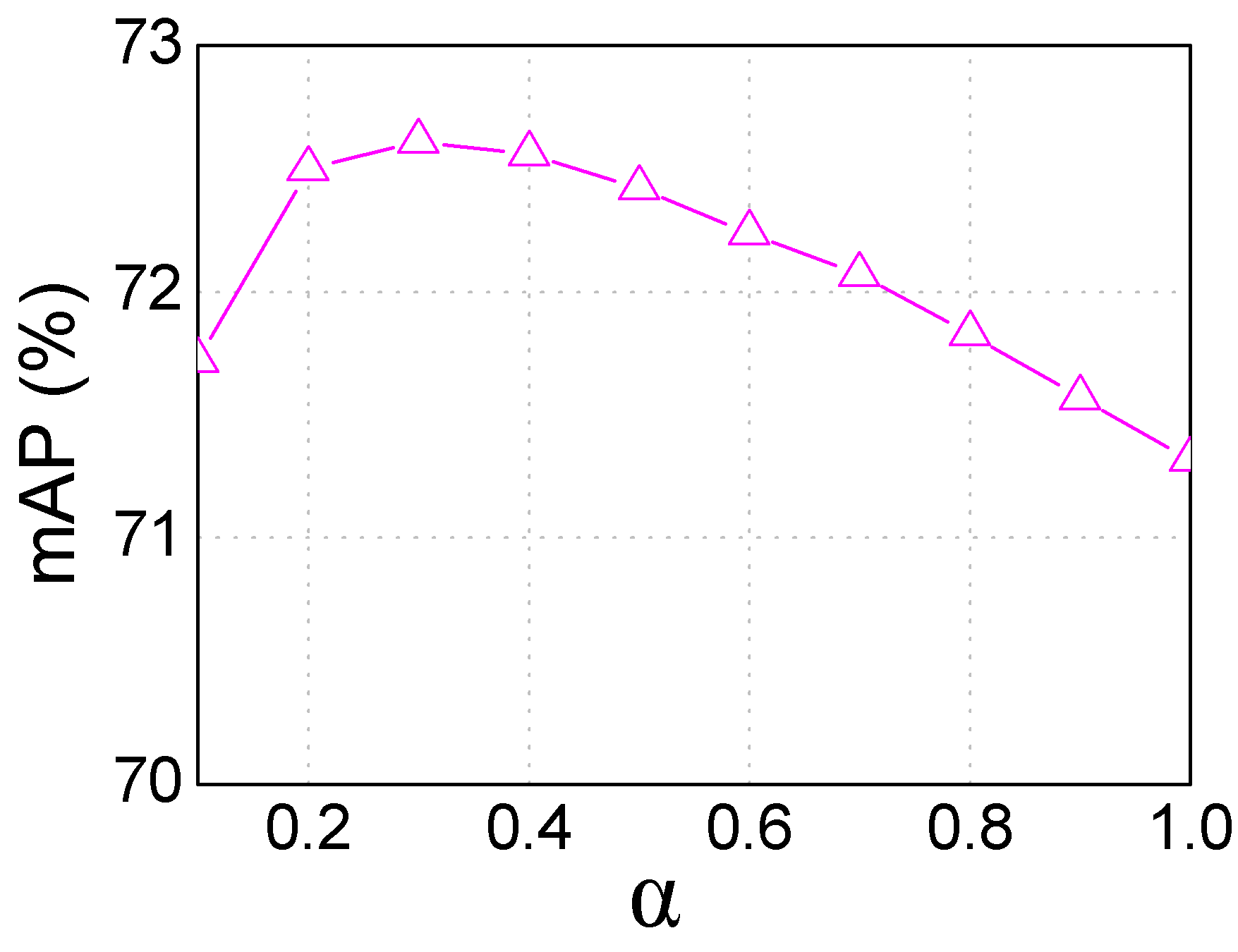}}\\
\multicolumn {2}{c}{(a) $L$} & \multicolumn {2}{c}{(b) $\alpha$}\\
\\
\bmvaHangBox{\includegraphics[width=2.85cm]{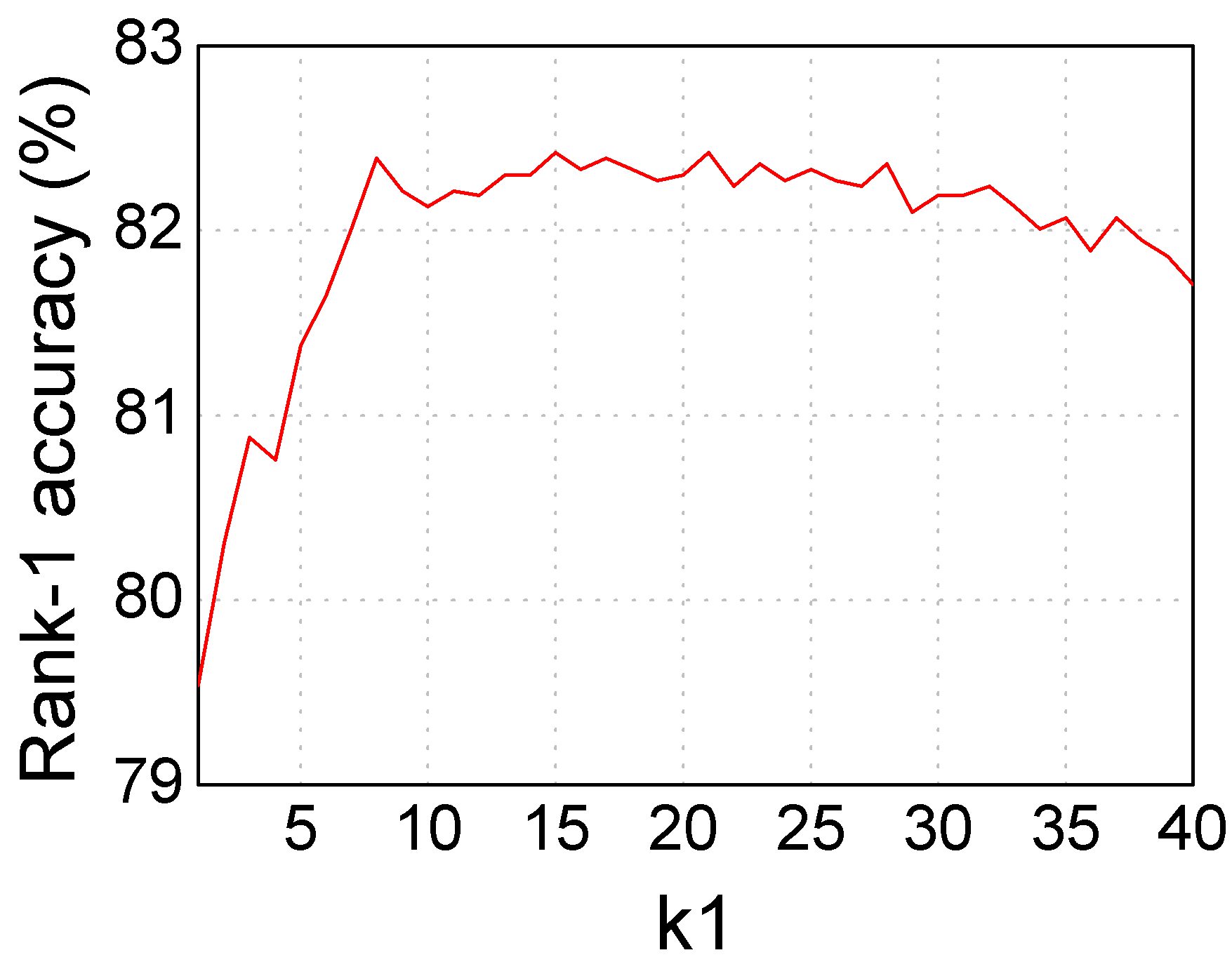}}&
\bmvaHangBox{\includegraphics[width=2.85cm]{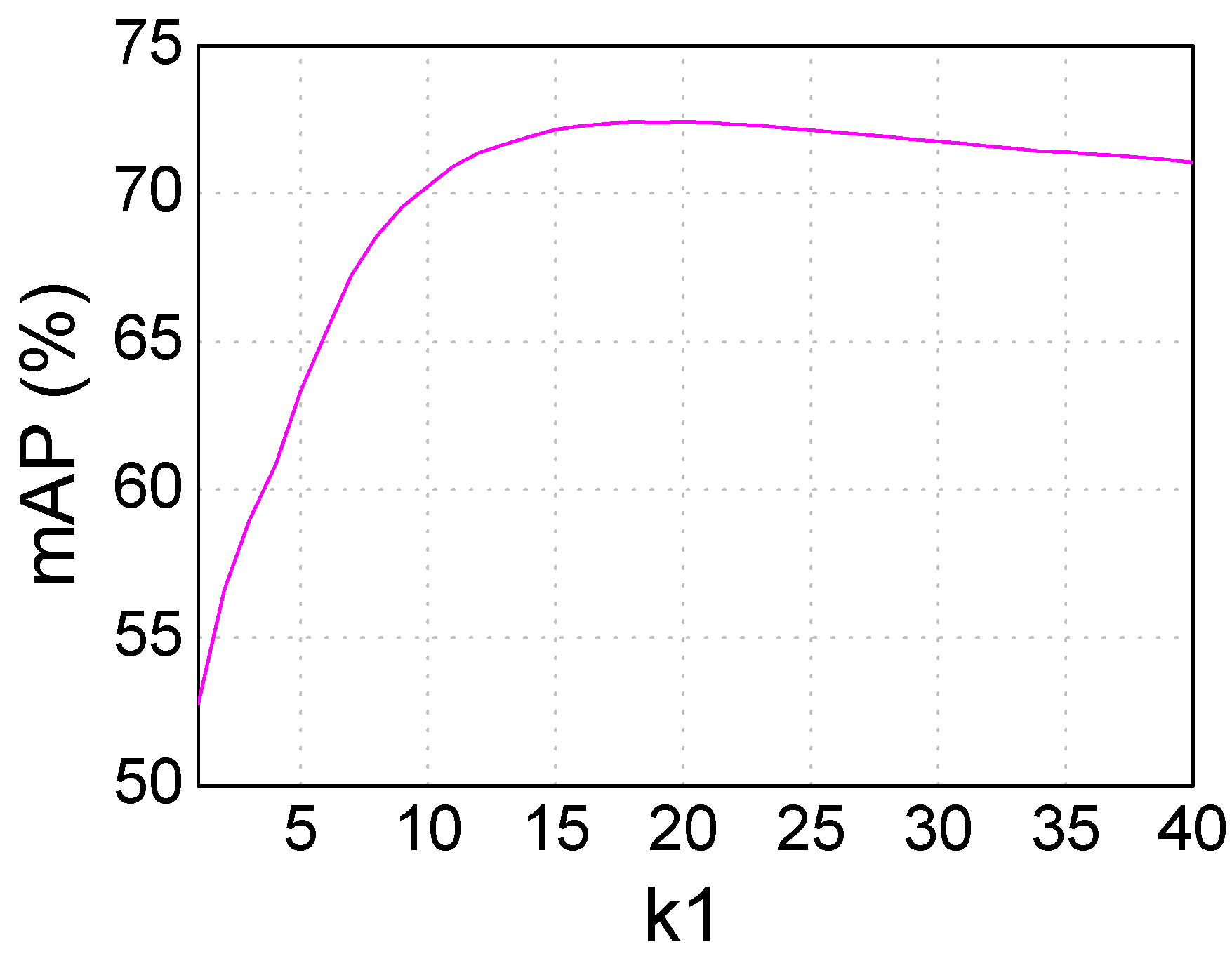}}&
\bmvaHangBox{\includegraphics[width=2.85cm]{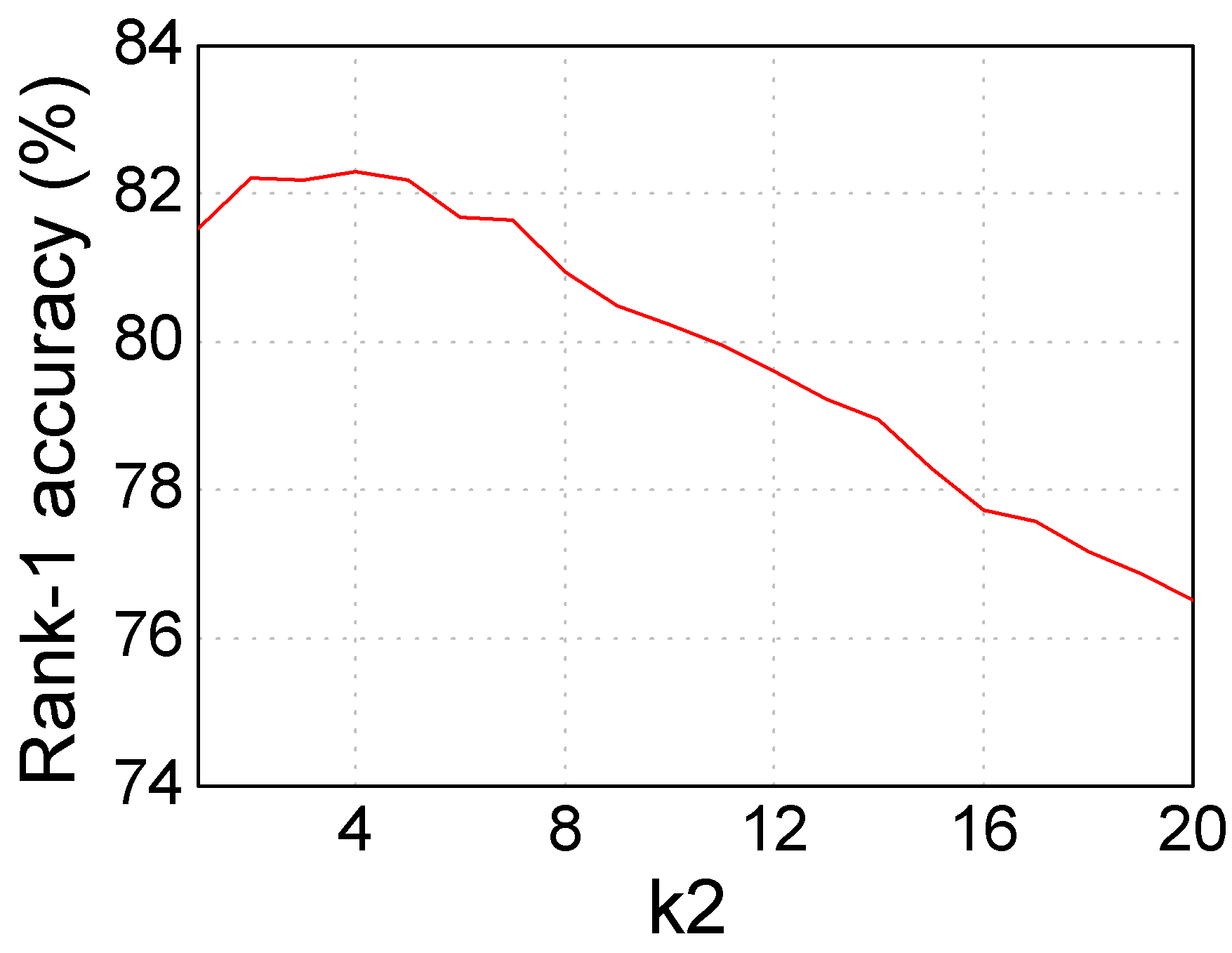}}&
\bmvaHangBox{\includegraphics[width=2.85cm]{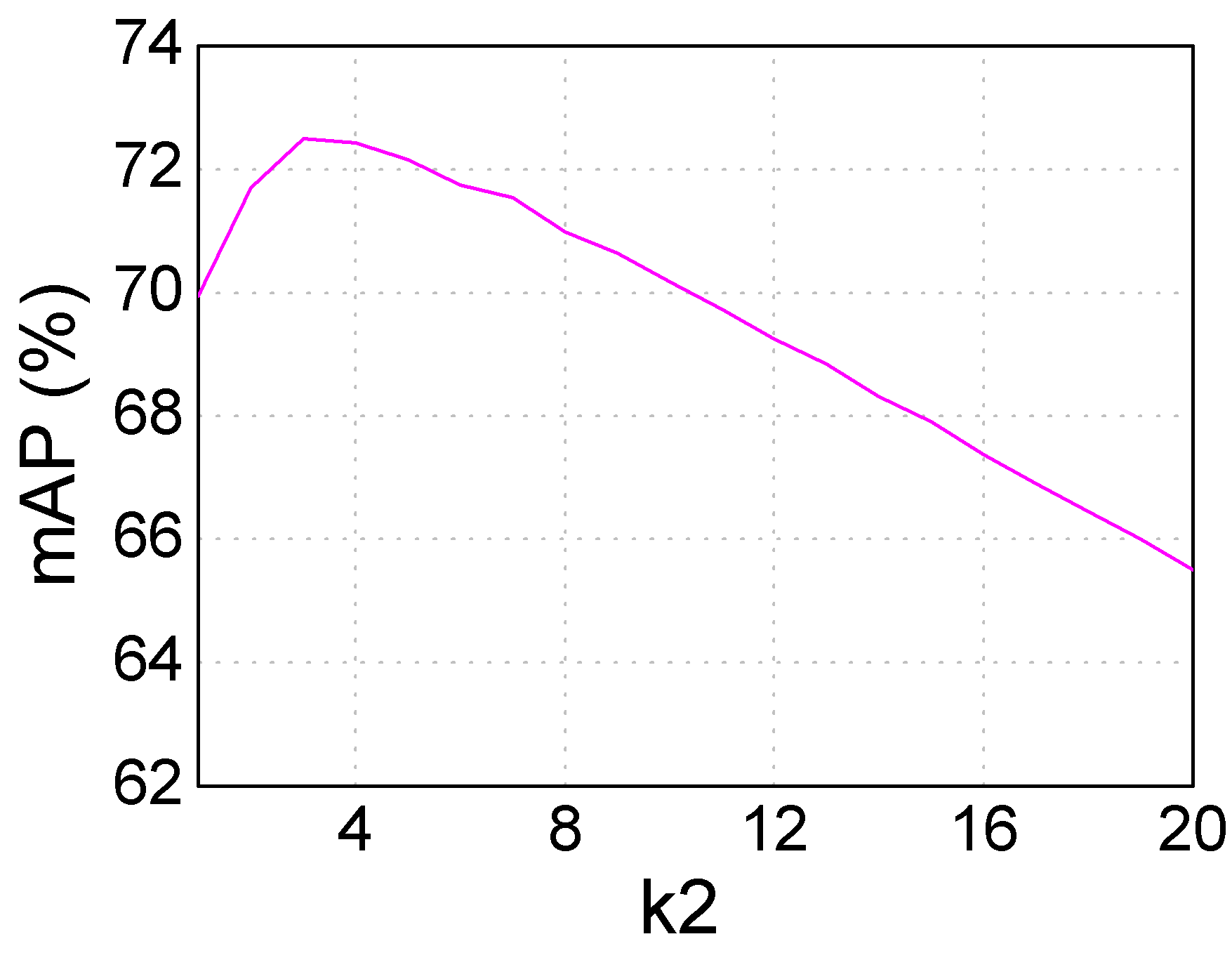}}\\
\multicolumn {2}{c}{(c) $k_1$} & \multicolumn {2}{c}{(d)  $k_2$}
\end{tabular}
\end{center}
\caption{The impact of different parameters on re-ID performance on Market-1501 dataset}
\label{fig:experiment_parameter}
\end{figure}

In this subsection, we experimentally analyze the impact of several parameters in our algorithm ($L$, $\alpha$, $k_1$ and $k_2$) on the re-ID performance. The baseline is based on the Euclidean metric of 2,048-dim ResNet features. We evaluate the proposed re-ranking approach on Market-1501 dataset under different parameter settings.

\vspace{+1ex}\noindent\textbf{The impact of $L$} is shown in Figure~\ref{fig:experiment_parameter} (a). With the increase of $L$, the rank-1 accuracy first rises and then fluctuates. When considering re-ID as a retrieval problem, we are more concerned with the mAP, which first rises as $L$ increases, and then descends after reaching a peak around $L$=10. There is a trade-off on the number of sub-features. A small $L$ may result in insufficient diversity introduced by feature division. On the other hand, a large $L$ leads to a relatively low-dimensional sub-feature (\eg, for a 2,048-dim ResNet feature, when $L$=32, the dimension of sub-feature is 64) with weak discriminating power, accordingly impairing the performance of the fused vector.

\vspace{+1ex}\noindent\textbf{The impact of $\alpha$} is given in Figure~\ref{fig:experiment_parameter} (b). As $\alpha$ increases, the rank-1 accuracy follows a slowly declining trend in fluctuation, while the mAP first rises and then drops steadily. The optimal point is around $\alpha=0.4$. In particular, when $\alpha$ is set to 1, the fuzzy aggregation operator in Eq.~\eqref{eq:fusion} turns into the arithmetic mean. As shown in Figure~\ref{fig:experiment_parameter} (b), the proposed fusion operator consistently outperforms the arithmetic mean.

\vspace{+1ex}\noindent\textbf{The impact of $k_1$} is charted in Figure~\ref{fig:experiment_parameter} (c). We fix $k_2$ to 4 and vary $k_1$. With the growth of $k_1$, the overall changing trends of the rank-1 accuracy and mAP are similar, that is, first stepping up and then initiating a slow descent. Recall that $k_1$ is the number of nearest neighbors whose similarities are encoded in the new feature. A large $k_1$ will encode superfluous similarities from the nearest neighbors and hence deteriorate performance.

\vspace{+1ex}\noindent\textbf{The impact of $k_2$} is presented in Figure~\ref{fig:experiment_parameter} (d). We fix $k_1$ to 20 and vary $k_2$. As $k_2$ increases, the rank-1 accuracy and mAP first rise and then decline steadily after reaching the peak at $k_2$=4. When $k_2$ is large, the neighbor enhancement aggregates too excessive similarities from the $k_1$-nearest neighbors of the probe's $k_2$-nearest neighbors and consequently degrade performance. Thus, $k_2$ is generally set smaller than $k_1$ for the closer neighborhood required by the neighbor enhancement in Eq.~\eqref{eq:enhance}.

\subsection{Further Evaluations}
\vspace{+1ex}\noindent\textbf{Contributions of components.}~We further conduct experiments on the Market-1501 dataset to assess the impact of each individual component. As shown in Table~\ref{tab:component_evalu},  (1) the ``Divide and Fuse'' scheme achieves 2.32\% mAP improvement; and (2) the iterative scheme improves mAP from 71.22\% to 72.42\% with slight decrease in Rank-1. The latter improvement is owing to the fact that the intermediate distance can somehow compensate the initial distance. Then we evaluate on different iterations. By tuning $\lambda$, more iterations only bring marginal improvements. Considering the increased CPU overhead, iterating twice is generally an advisable setting to achieve satisfactory performance.
\begin{table}
\begin{center}
\begin{tabular}{|l|c|c|}
\hline
\multicolumn{1}{|c|}{Components} & Rank-1 & mAP\\
\hline\hline
Without ``Divide and Fuse'' scheme & 80.91 & 70.10 \\
Without iterative encoding & 82.42 & 71.22 \\
Our full model & 82.30 & 72.42 \\
\hline
\end{tabular}
\end{center}
\caption{Evaluation on individual components}
\label{tab:component_evalu}
\end{table}

\vspace{+1ex}\noindent\textbf{Running cost.}~All the experiments are performed on a server with 3.20 GHz CPU and 64 GB memory. It is worth mentioning that our method is very efficient thanks to the inverted index for sparse vectors. When L=11 and iteration=2, it costs about 0.12 seconds per query to match 19,732 testing images on the Market1501 dataset.

\vspace{+1ex}\noindent\textbf{Applied to other tasks.}~To further evaluate our method, we apply it to the image retrieval task and evaluate on the Holidays dataset. The deep feature is extracted by a CNN model as the baseline. The experimental results show our method outperforms other competitive re-ranking algorithms. For example, the baseline and SCA~\cite{bai2016sparse} achieve mAP 81.23\% and 82.88\%, respectively, both lower than 84.49\% achieved by our DaF algorithm. It can be expected that the proposed method holds great potential applications in other tasks.

\section{Conclusion}
\label{sec:conclusion}
In this paper, we propose a totally unsupervised person re-ID re-ranking framework, which provides a simple yet effective means of exploiting the diverse information embedded in a high-dimensional feature. We also present how to iteratively encode the contextual information from three aspects so as to fully utilize neighborhood relations. Our method can serve as a post-processing tool to boost the performance upon most initial ranking lists. Experimental results demonstrate that our approach can consistently improve the re-ID accuracy and outperform other re-ranking methods. In the future, we will investigate the effective way of fusing sub-feature information from multiple features.

\bibliography{egbib}
\end{document}